\documentclass{egpubl}

 \ShortPresentation    
 \electronicVersion 

\ifpdf \usepackage[pdftex]{graphicx} \pdfcompresslevel=9
\else \usepackage[dvips]{graphicx} \fi

\PrintedOrElectronic

\usepackage{t1enc,dfadobe}

\usepackage{egweblnk}
\usepackage{cite}

\title[Towards automatic initialization of registration algorithms]%
      {Towards automatic initialization of registration algorithms using simulated endoscopy images}

\author[A. Sinha \& M. Ishii \& R.\,H. Taylor \& G.\,D. Hager \& A. Reiter]
{\parbox{\textwidth}{\centering Ayushi Sinha$^{1}$, Masaru Ishii$^{2}$, Russell H. Taylor$^{1}$, Gregory D. Hager$^{1}$ 
and Austin Reiter$^{1}$
        }
        \\
{\parbox{\textwidth}{\centering $^1$The Johns Hopkins University, Baltimore, MD, USA\\
        $^2$Johns Hopkins Medical Institutions, Baltimore, MD, USA
       }
}
}

\begin{document}


\maketitle
\begin{abstract}
Registering images from different modalities is an active area of research in computer aided medical interventions. Several registration algorithms have been developed, many of which achieve high accuracy. However, these results are dependent on many factors, including the quality of the extracted features or segmentations being registered as well as the initial alignment. Although, several methods have been developed towards improving segmentation algorithms and automating the segmentation process, few automatic initialization algorithms have been explored. In many cases, the initial alignment from which a registration is initiated is performed manually, which interferes with the clinical workflow. Our aim is to use scene classification in endoscopic procedures to achieve coarse alignment of the endoscope and a preoperative image of the anatomy. In this paper, we show using simulated scenes that a neural network can predict the region of anatomy (with respect to a preoperative image) that the endoscope is located in by observing a single endoscopic video frame. With limited training and without any hyperparameter tuning, our method achieves an accuracy of 76.53\,($\pm$\,1.19)\,\%.
There are several avenues for improvement, making this a promising direction of research. Code is available at \url{https://github.com/AyushiSinha/AutoInitialization}.


\end{abstract}
\section{Introduction}

Several surgical procedures, especially minimally invasive surgeries (MIS), require registration or alignment between a preoperative image and an intraoperative image in order to provide additional knowledge available from preoperative images~\cite{Mezger13}. For MIS through the nasal cavity, these modalities are generally computed tomography (CT) scans and endoscopic video. Registration provides contextual cues to augment the limited information provided by endoscopic video. This allows surgeons to locate their endoscope and tools in relation to critical anatomical structures visible in CT scans, making the surgery safer~\cite{Senior09}. Registration also enables navigation during surgery, allowing surgeons to additionally track the endoscope and tools within the CT coordinate frame. High accuracy in these registrations is crucial because of the proximity of critical anatomical structures, like the carotid arteries, optic nerves, eyes, brain, etc., to the nasal cavity region~\cite{Tao99}. However, registration algorithms generally require a coarse initial alignment to bring the two modalities \emph{close enough} before launching the registration, 
and registration accuracy is dependent on the quality of this initial alignment. Often, this initial alignment is performed manually~\cite{Mori02, Higgins03, Leonard16}. 
This can be tedious and interferes with the clinical workflow. Accurate automatic initialization of registration algorithms is one of the major components required to fully automate surgical navigation, making this an important area of research.

Many previous methods have attempted to automatically initialize registration algorithms for different applications. Methods have been presented that align features from the two modalities to a common coordinate frame before computing the final initialization~\cite{Foroughi08} or that align the principal modes of variation to find the initial coarse alignment~\cite{Bricq18}. In video based registration algorithms, several methods rely on finding canonical landmarks or other features in both modalities~\cite{Miao12, Robu18}. These features can be specific to certain anatomical structures and, therefore, hard to generalize. Many of these methods require several preprocessing steps which can be time-consuming as well as prone to noise and outliers.

We present a method that takes as input a frame of endoscopic video and directly predicts the region in which the endoscope must be located in order to generate the view seen in the video frame. Our preliminary results are on simulated data and show that our region classifier is able to identify the region that the endoscope is in with an accuracy of $76.53$\,($\pm\,1.19$)\,\%. We are currently working on vastly expanding the dataset and hope to extend this method to real endoscopy video.

\section{Methods}

We approach the problem of camera localization by trying to learn the types of views observed from cameras located in different regions of the nasal cavity. However, it is difficult to generate reliable ground truth in large numbers of in-vivo endoscopy images because the exact location of the camera is unknown. Therefore, we generate labeled data in simulation where the exact ground truth is known in order to learn the types of views that are observed from different camera poses. In the following sections, we explain how the simulated data is generated and how the camera location is learned.

\subsection{Data generation}

\begin{figure}[tb]
   \centering \includegraphics[width=1\linewidth]{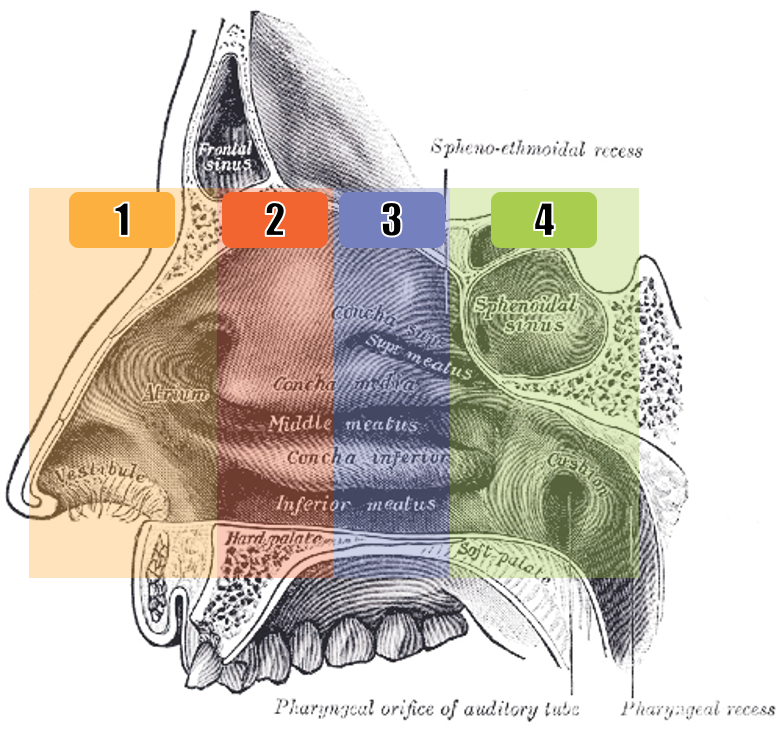}
   \caption{\label{fig:regions}
     The $4$ regions of the nasal cavity: images acquired from cameras located in these different regions of the nasal cavity contain different salient features which we want to learn.}
\end{figure}

\begin{figure}[tb]
   \centering \includegraphics[width=0.939\linewidth]{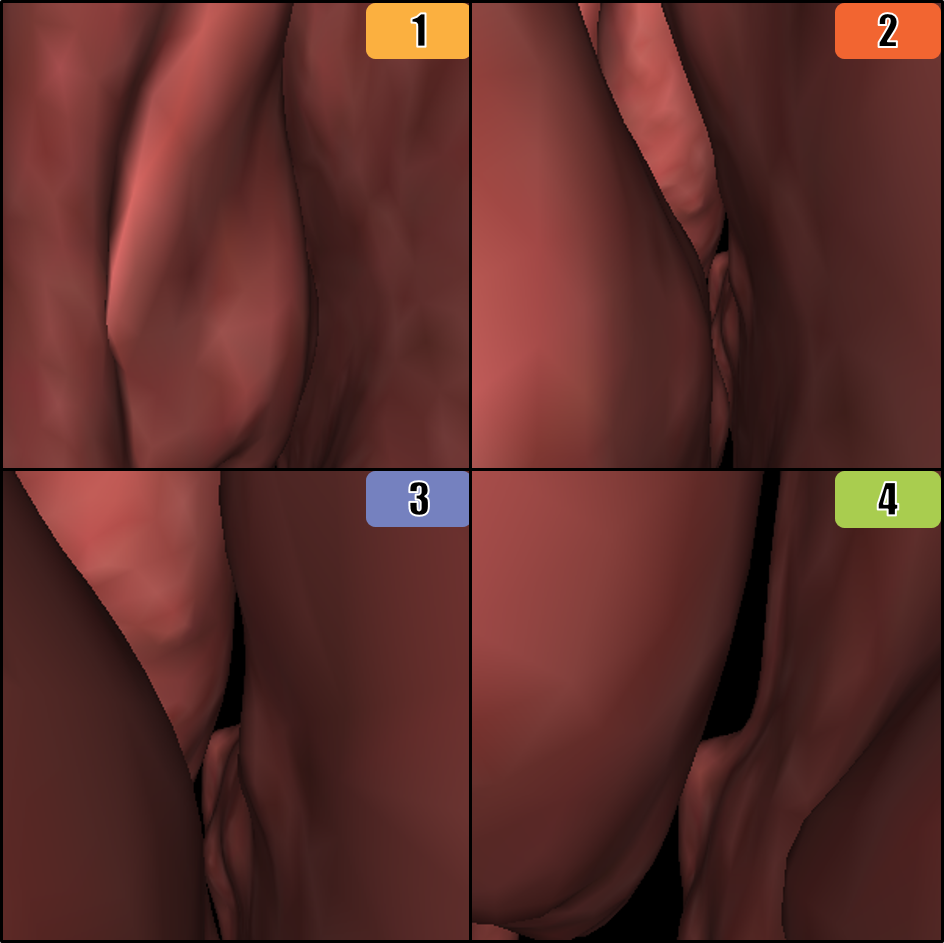}
   \caption{\label{fig:examples}
    Examples of the different scenes observed from cameras located in the $4$ different regions of the nasal cavity.}
\end{figure}

Our simulated dataset is generated in OpenGL using a textured mesh of the nasal cavity. The mesh is a mean mesh built from a publicly available dataset~\cite{Beichel15, Bosch15, Clark13, Fedorov16} of $53$ CTs which were automatically segmented~\cite{Sinha17}. This mean mesh is textured using an image generated from in-vivo nasal endoscopy video. A camera and a single light source are co-located and steered through this textured mesh environment. At each camera location, the camera is rotated slightly to observe $10$ random views. The images rendered from these views as well as the corresponding camera poses are saved. The camera poses are grouped into $4$ classes based on the region that the camera is located in (Figure~\ref{fig:regions}). Examples of rendered images from each class are shown in Figure~\ref{fig:examples}.

\subsection{Classification}

We trained a $2$ layer convolutional neural network~\cite{LeCun15} with cross entropy loss to learn the classes associated with the rendered images. Details about our network are fully specified in our code, available at \url{https://github.com/AyushiSinha/AutoInitialization}. Our training dataset contained $1270$ images, and we tested on a separate set of $150$ images. Since we are not performing any hyperparameter tuning, we do not have a validation set. We trained and tested our network $5$ times, each time training for a total of $30$ epochs with a learning rate and weight decay factor of $1e^{-3}$. We report the average error for evaluation.


\section{Results and discussion}
\begin{figure}[tb]
   \centering \includegraphics[width=1\linewidth]{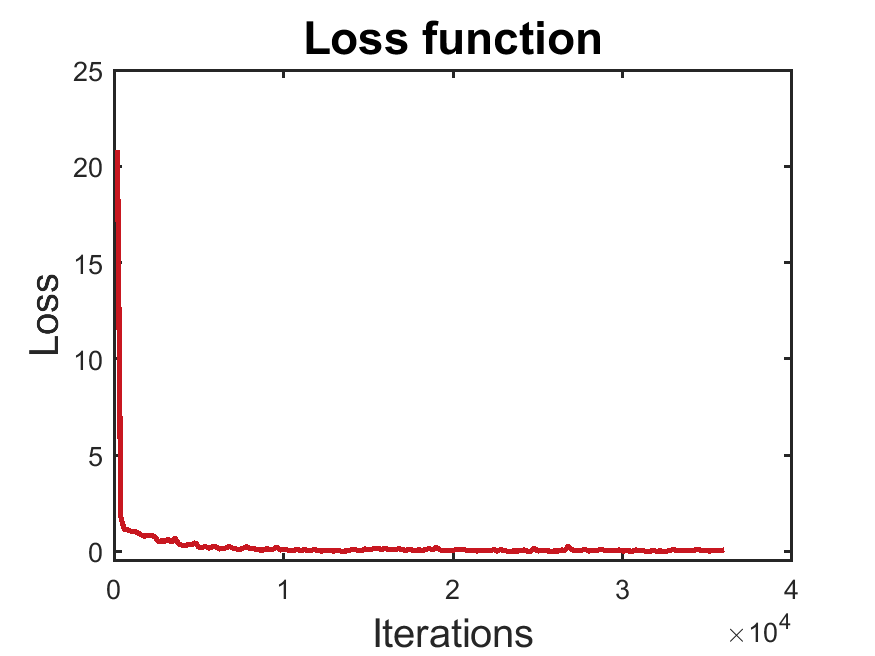}\vspace{3.5mm}
   \centering \includegraphics[width=1\linewidth]{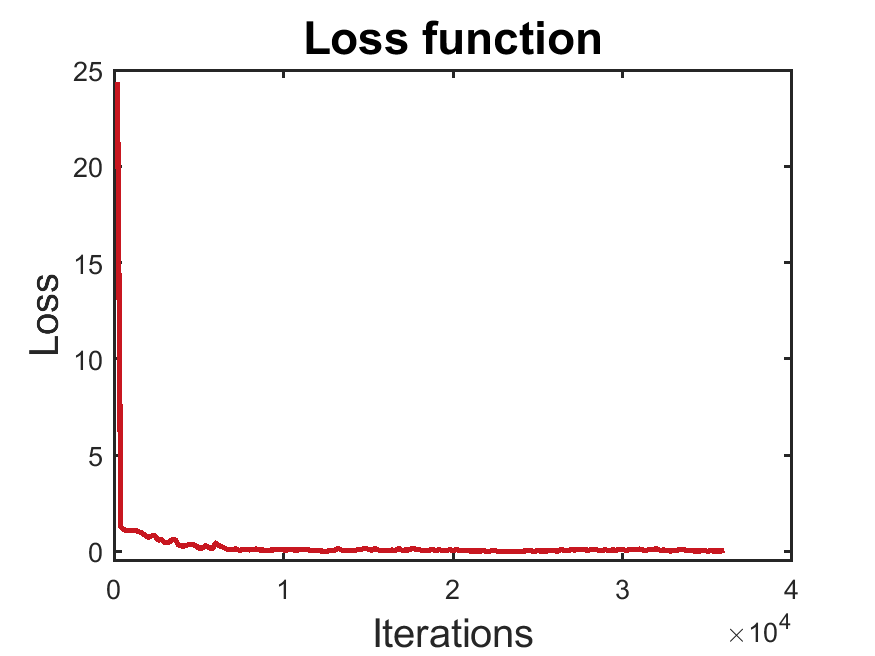}
   \caption{\label{fig:loss}
    Evolution of the loss function from run $1$ (above) and run $3$ (below): the loss function converges quickly and remains stable for all runs.}
\end{figure}

\begin{figure}[tb]
   \centering \vspace{1mm}\includegraphics[width=0.93\linewidth]{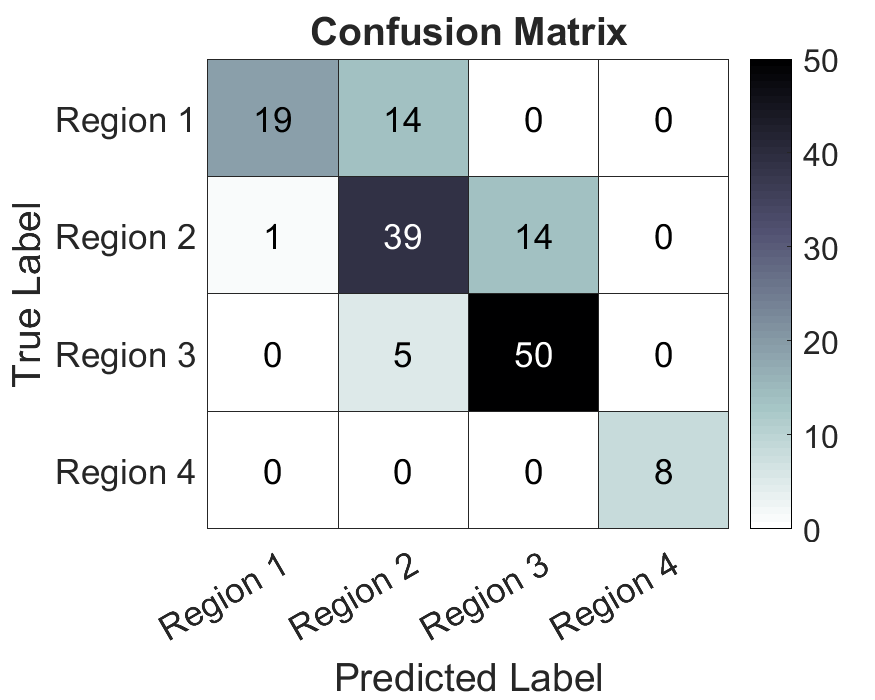}\vspace{3.25mm}
   \centering \includegraphics[width=0.93\linewidth]{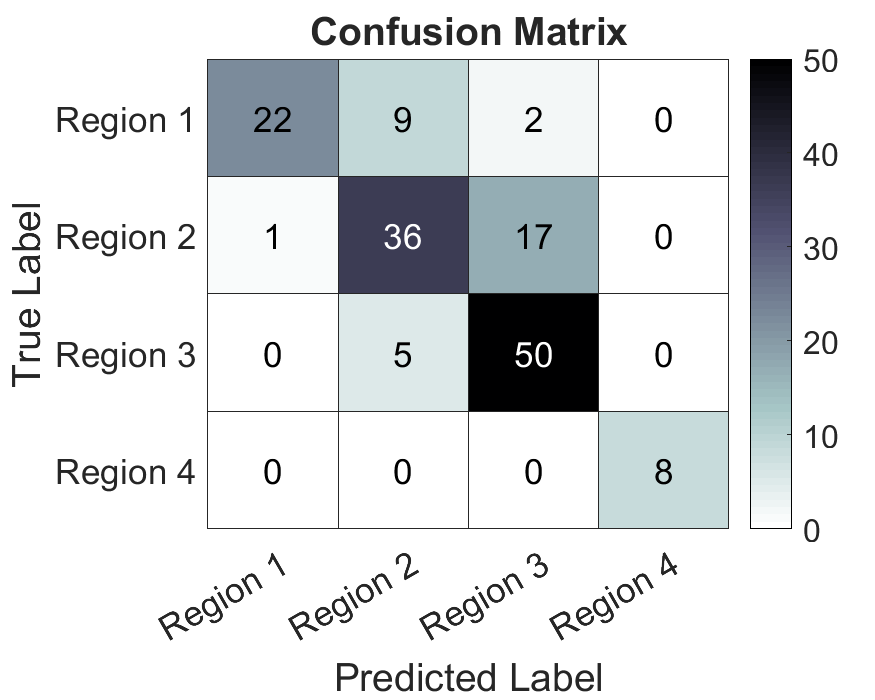}
   \caption{\label{fig:cm}
    Confusion matrix for run $1$ (above) and run $3$ (below): most labels are classified correctly and errors in classification are generally with neighboring classes.} 
\end{figure}

We were able to train our classifier to learn the class associated with each image (Figure~\ref{fig:loss}). The images in our test dataset were classified with a mean accuracy of $76.53$\,($\pm\,1.19$)\,\% over $5$ runs. Further, the errors in classification are almost always with neighboring classes (Figure~\ref{fig:cm}). That is, images in \emph{Region 2} are sometimes misclassified as belonging to \emph{Region 1} or \emph{Region 3}, but never as belonging to \emph{Region 4}. This is reasonable given that images rendered from endoscopes located at the border of the specified regions (Figure~\ref{fig:regions}) are likely to be similar in appearance but have different associated labels. A larger training dataset and tuning the hyperparameters can help improve such errors.

These preliminary results are promising given the limited size of our current training dataset and that no hyperparameter tuning has been performed. We are currently working on expanding our training dataset in order to generalize better to different scenarios. Different images will be used to texture the mean mesh to simulate different patients and different lighting conditions will be added to simulate different endoscopes. Further, we hope to use transfer learning to extend this classification to in-vivo endoscopy video and evaluate whether these classifications are accurate enough to result in reliable final registrations.

We will continue to attempt to associate the simulated and in-vivo endoscopy images to regions in the mean mesh because we expect all nasal passages to contain some number of common salient features. If a patient specific CT is segmented such that the patient mesh is in correspondence with the mean mesh, 
then the associated region in the mean mesh can be easily transfered to the patient mesh. If the two meshes are not in correspondence, then a registration between the two might also require manual initialization. However, since CT scans are available before the endoscopic procedure, the initialization and registration can be performed offline without interfering with the surgical workflow. Once a region in the patient mesh is determined, registrations between the patient CT and features extracted from patient endoscopic video can be initiated from multiple poses within the specified region. The registration with the lowest error or highest stability~\cite{Leonard18} can be chosen as the final registration. If a patient CT is not available, then recent methods are able to deformably register features from patient endoscopic video to a statistically mean mesh~\cite{Sinha18}. In this case, registrations can be initiated directly within the region in the mean mesh determined by our classifier. Further, with a much larger dataset, we hope to additionally be able to estimate the camera pose that would generate a rendered image and use the estimated pose to initialize the registration.

\section{Conclusions}

This work shows preliminary results towards building a closed loop surgical navigation pipeline that does not require any user interaction. Our results show that it is possible to reliably classify the region that an endoscope is located in by simply observing a single simulated endoscopic image. We hope to further improve our accuracy and achieve such classifications in in-vivo endoscopic images. Assuming that features from video and preoperative images can be extracted automatically, further work in this area holds the potential for fully automated registrations and, consequently, seamless navigation during clinical endoscopic explorations as well as during endoscopic surgery.


\section*{Acknowledgment}

This work was funded by NIH R01-EB015530, the Provost's Postdoctoral Fellowship at the Johns Hopkins University, fellowship support from Intuitive Surgical, Inc., and the Johns Hopkins University internal funds.



\bibliographystyle{eg-alpha-doi}

\bibliography{egbibsample}

%
%

\end{document}